\documentclass{article}    %定义文章类型为article
\usepackage{spconf,amsmath,graphicx,hyperref,cite,helvet,enumitem,courier,array}    %引入所需要的宏包(扩展功能)
\usepackage{booktabs,xcolor,amssymb,caption, bm, multirow, makecell, soul}    % for \toprule, \midrule, \bottomrule in tables
\usepackage[export]{adjustbox} % 关键宏包
\usepackage{etoolbox}
\usepackage{fontawesome}

\definecolor{orangehighlighter}{RGB}{255,165,0} % 橙色高亮
\definecolor{yellowhighlighter}{RGB}{255,255,0} % 黄色高亮
\title{FAST-GS: Frequency Aware Space-time Gaussian Splatting for Photorealistic Dynamic Novel View Synthesis}
\name{%
\begin{tabular}{@{}c@{}}
Zhengyang Zhang$^{1\bigstar}$, Ziyu Lu$^{1\bigstar}$, PengCheng Li$^{1}$, Hongbo Duan$^{1}$,\\
Yi Liu$^{1}$, Pengting Luo$^{2}$, Peiyu Zhuang$^{2*}$, Xinghui Li$^{1*}$, Shaohua Ma$^{1*}$
\end{tabular}%
\thanks{${}^{\bigstar}$\textit{These authors contributed equally to this work.}}%
\thanks{${}^{*}$\textit{Corresponding author. E-mail(s): 
\href{mailto:ma.shaohua@sz.tsinghua.edu.cn}{ma.shaohua@sz.tsinghua.edu.cn},
\href{mailto:li.xinghui@sz.tsinghua.edu.cn}{li.xinghui@sz.tsinghua.edu.cn},
\href{mailto:zhuangpeiyu2022@163.com}{zhuangpeiyu2022@163.com}}}%
}
 \address{$^{1}$Shenzhen International Graduate School, Tsinghua University, Shenzhen, China \\ $^{2}$ Central Media Technology Institute, Huawei}

\begin{document}
\topmargin=0mm
\ninept
\maketitle

\begin{abstract}
4D Gaussian Splatting (4DGS) excels in dynamic 3D reconstruction and real-time novel view synthesis via efficient 4D Gaussian representations and parallelizable rendering. However, existing 4DGS approaches rely on a single polynomial to model motion—this limits performance in complex dynamic scenes where high-frequency motion components are prevalent, and fails to ensure long-term stability due to cumulative trajectory drift. To address these issues, we propose a Fourier Motion Modeling module: this paradigm decomposes motion into frequency-based sinusoidal components, capturing both low-frequency global trajectories and high-frequency local details to model complex motion patterns accurately. It retains 4DGS’s real-time rendering capability while improving complex motion fitting and long-term coherence. Additionally, we integrate a motion-aware regularization strategy into the loss function: it uses frequency-dependent weights to suppress high-frequency jitter while preserving low-frequency motion coherence. Extensive experiments on N3V and Google Immersive datasets from multiple scenarios demonstrate the effectiveness of our method.
\end{abstract}

%第一章：INTRODUCTION
\section{Introduction}
\label{sec:intro}

Accurate, photorealistic rendering of dynamic 3D scenes is critical for immersive media (VR/AR), sports broadcasting, and film production. A core challenge in this field is balancing high fidelity (capturing complex motion) and temporal coherence (maintaining stability in long sequences)—existing methods struggle to address both simultaneously.

Recent years have witnessed remarkable progress in the field of neural rendering. Neural Radiance Fields (NeRF)\cite{mildenhall2021nerf} and their dynamic 
extensions\cite{barron2023zip,chen2022tensorf,chen2023neurbf,neff2021donerf,piala2021terminerf,li2021neulf} have enabled high-quality novel view synthesis 
and dynamic scene reconstruction through neural networks and volumetric rendering. However, their substantial computational cost has hindered real-time 
high-definition rendering of complex dynamic scenes. More recently, 3D Gaussian Splatting (3DGS)\cite{kerbl20233d} has attracted considerable attention 
due to its capacity for high-quality real-time rendering in unbounded static scenes, and is increasingly supplanting NeRF as the mainstream backbone for 
3D reconstruction models. By employing an explicit Gaussian representation, 3DGS forgoes the generalization ability inherent in NeRF's continuous neural 
field representation, yet achieves real-time performance without sacrificing visual quality. For dynamic scenes, Dynamic 3D Gaussians\cite{luiten2024Dynamic3D} 
facilitate the reconstruction of multi-view dynamic scenes by modeling per-frame Gaussian positions and rotations independently. Nevertheless, this method 
encounters challenges of escalating complexity and excessive memory consumption when processing long sequences.

%图1:4dgs切片流程
\begin{figure}[!t]

\begin{minipage}[b]{1.0\linewidth}
  \centering
  \centerline{\includegraphics[width=8.5cm]{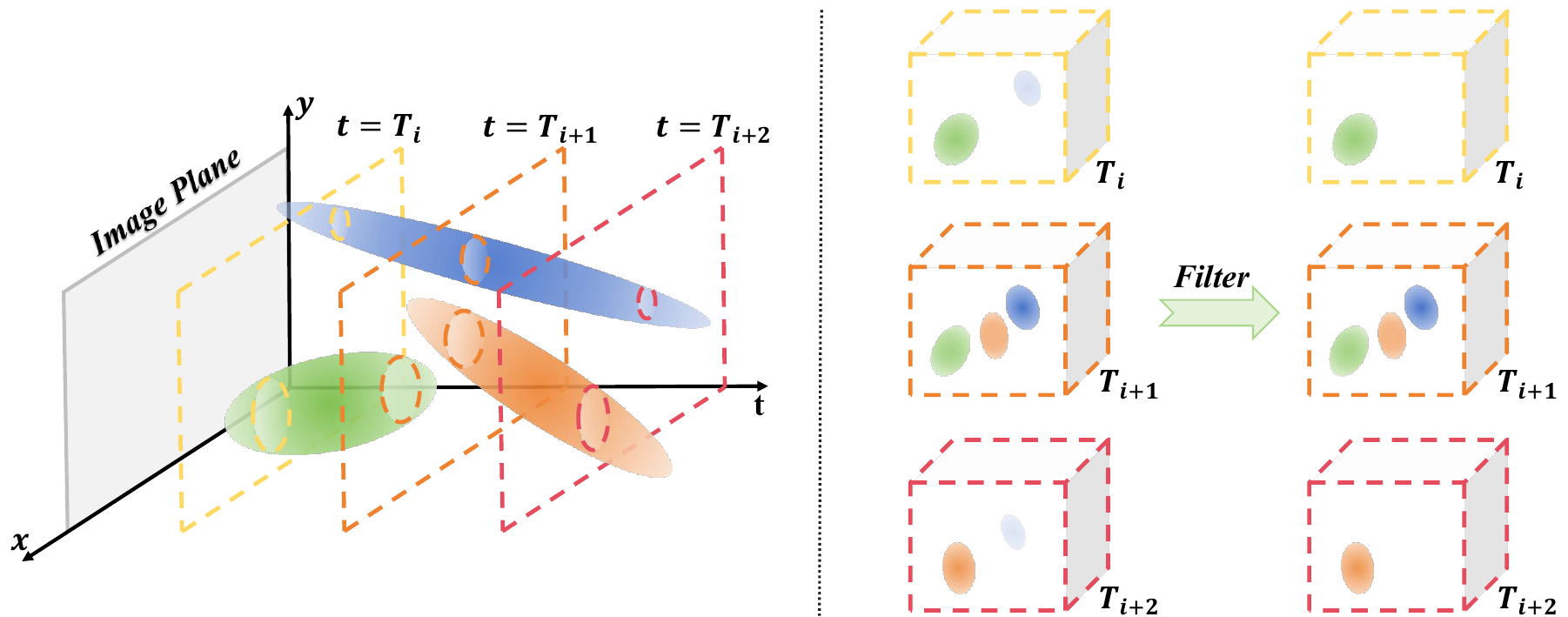}}
%  \vspace{2.0cm}
 % \centerline{(a) Result 1}\medskip
\end{minipage}
\caption{\textbf{A simplified illustration of temporal slicing within a 4DGS.} The 4D Gaussian is conceptualized as a hypercylinder in 4D space. For a given time query, 
corresponding 3D Gaussian ellipsoids are extracted. Their color depth indicates temporal opacity; ellipsoids falling below a predefined opacity threshold are filtered from rendering.}
\label{fig:1}
\end{figure}

%图2:pipeline
\begin{figure*}[!t]
\centering
\includegraphics[width=0.9\linewidth]{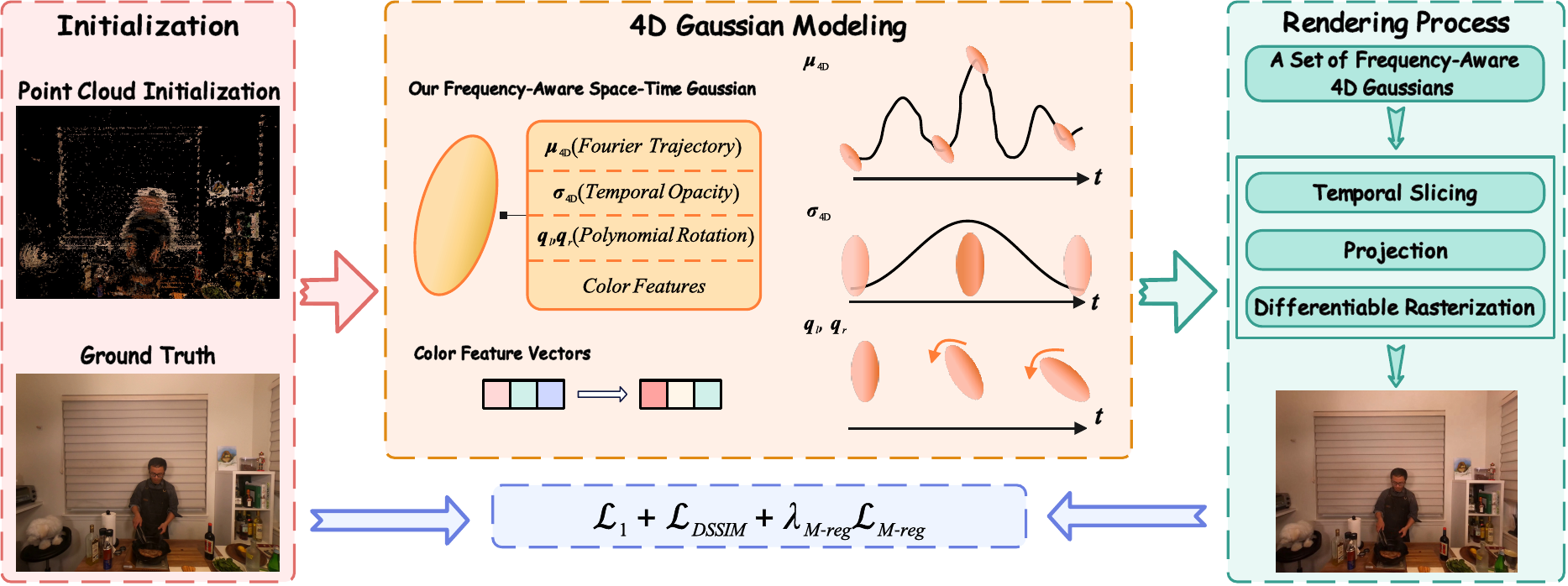}
%  \vspace{2.0cm}
 % \centerline{(a) Result 1}\medskip
\caption{\textbf{Pipeline of FAST-GS.} Given a scene input, after point cloud initialization, our FAST-GS further incorporates Fourier-based motion, 
temporal opacity, polynomial rotation, and time-dependent features. Following 4D Gaussian modeling, our rendering process\cite{4dgsRealtime} consists of three stages: temporal slicing, projection, and differentiable 
rasterization, ultimately producing the novel view rendering. The output image is optimized using a loss function enhanced with motion regularization.}
\label{fig:2}
\end{figure*}

Building upon this foundation, 4D Gaussian Splatting (4DGS)\cite{4dgsdeform,4dgsRealtime,4drotor} models time-varying scenes as four-dimensional spatiotemporal Gaussian hyperprisms:\cite{4dgsdeform,yang2024deformable3D,attal2023hyperreel} motion using 6-DoF trajectories or deformation fields
to learn inter-frame Gaussian transformations,\cite{4dgsRealtime,4drotor,lee2024fully} adopt 4D Gaussian primitives to integrate spatio-temporal structures and corresponding features for real-time rendering
of dynamic content. As shown in Fig.\ref{fig:1}, 
at each time step, the 4D Gaussian is sliced into 3D Gaussians with time-varying position and transparency attributes. Transient content (such as appearing or disappearing objects) 
is filtered out via thresholding, and the remaining Gaussians are rasterised and projected onto a two-dimensional screen. By directly optimising the 4D Gaussian set, 
4DGS effectively represents both static and dynamic scene components simultaneously, achieving high-fidelity modelling. However, existing 4DGS approaches predominantly employ a 
single polynomial function\cite{4dgsRealtime,4drotor,li2024spacetime} to parameterise the spatiotemporal motion of 3D Gaussians — this design has two critical flaws:
\begin{enumerate}[label=(\arabic*)]
    \item \textit{High-frequency motion fitting failure}: Polynomials suppress high-frequency components (e.g., flame's flickering), leading to blurry rendering of fast, local motion.
    \item \textit{Long-term trajectory drift}: Polynomial parameters accumulate errors in long sequences, causing temporal incoherence.
\end{enumerate}
To solve these flaws, we introduce FAST-GS (Frequency Aware Space-Time Gaussian Splatting). Its core insight is that Fourier series — a classic signal processing tool -  naturally decomposes the spatiotemporal motion ${(x(t), y(t), z(t))}$ of Gaussian points into multi-frequency components, making it ideal for modeling dynamic motion. By replacing polynomial motion modeling with fourier decomposition and adding frequency-weighted regularization, this approach retains the real-time rendering benefits of 4DGS while addressing two key challenges: complex motion fitting and long-term stability.

% These guidelines include complete descriptions of the fonts, spacing, and
% related information for producing your proceedings manuscripts. Please follow
% them and if you have any questions, direct them to Conference Management
% Services, Inc.: Phone +1-979-846-6800 or email
% to \\\texttt{papers@2026.ieeeicassp.org}.

%第二章：METHOD
\section{METHOD}
%\label{sec:pagestyle}
\label{sec:method}

%\section{TYPE-STYLE AND FONTS}
%\label{sec:typestyle}

%图3:对比试验图
\begin{figure*}[!t]
\centering
\includegraphics[width=0.9\linewidth]{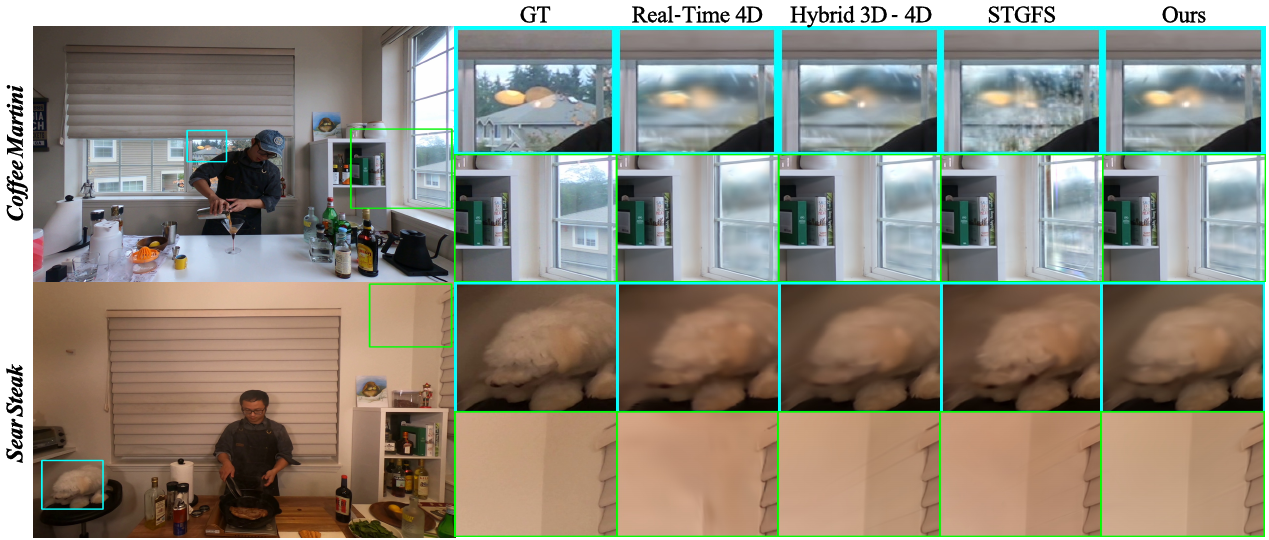}
%  \vspace{2.0cm}
 % \centerline{(a) Result 1}\medskip
\caption{\textbf{Qualitative comparison on the N3V dataset.} While most methods yield comparable results, our approach demonstrates superior performance in modeling details of challenging and complex regions or objects.
Zoom in for best viewing.}
\label{fig:3}
\end{figure*}

\subsection{\textit{Preliminary: 3D and 4D Gaussian Splatting}}
\label{ssec:subhead}
3D Gaussian Splatting (3DGS)\cite{kerbl20233d} represents scenes with a set of anisotropic 3D Gaussians. Each Gaussian is defined by a center ${\mu}$ and a covariance matrix ${\Sigma}$, 
which determine its spatial extent and orientation. The influence of a Gaussian at a 3D point $\mathbf{x}$ is given by:
\begin{equation}
  \label{eq:3dgaussian}
G(\mathbf{x})=\exp\left(-\frac{1}{2}(\mathbf{x}-{\mu})^{\top} {\Sigma}^{-1}(\mathbf{x}-{\mu})\right)
\end{equation}

The covariance is decomposed as $\Sigma = R S S^\top R^\top$, where $R$ is a rotation matrix and $S$ a scaling matrix. Rendering uses alpha compositing: the pixel color $C$ is computed by blending ordered Gaussians:
\begin{equation}
  \label{eq:colorrendering}
{C=\sum_{i\in\mathcal{N}}c_{i}\alpha_{i}\prod_{j=1}^{i-1}(1-\alpha_{j})}
\end{equation}
where $c_i$ and $\alpha_i$ denote the color and opacity of each Gaussian. This enables efficient and photorealistic rendering.

4D Gaussian Splatting (4DGS)\cite{4dgsRealtime} achieves this by augmenting 3D Gaussians with time-varying Gaussian parameters (rotation, position). The 4D rotation is constructed as a product of two quaternion-based matrices:$R = R_l R_r$,
where ${R}_l$ and ${R}_r$ are left and right rotation matrices defined by quaternions $(a,b,c,d)$ and $(p,q,r,s)$ respectively. The position and scale are given by:
%\begin{equation}
%  \label{eq:muxyz}
$\mu_{\textit{xyz}|t}=\mu_{1:3}+\Sigma_{1:3,4}\Sigma_{4,4}^{-1}(t-\mu_{t})$ and
%\end{equation}
%\begin{equation}
%  \label{eq:Sigmaxyz}
$\Sigma_{\textit{xyz}|t}=\Sigma_{1:3,1:3}-\Sigma_{1:3,4}\Sigma_{4,4}^{-1} \Sigma_{4,1:3}$.
%\end{equation}

These time-varying parameters are all modeled by polynomials, which is the root cause of the limitations of 4DGS: it cannot represent high-frequency variations and accumulates errors over time.

\subsection{\textit{Frequency Aware Space-Time Gaussians}}
\label{ssec:subhead}
To model complex, long-duration motion, we redesign 4DGS’s motion representation using Fourier series as shown in Fig.\ref{fig:2}. 
Specifically, for each 3D Gaussian point, its instantaneous position in Eq.(\ref{eq:3dgaussian}) \(\mathbf{\mu}(t) = [x(t), y(t), z(t)]^T\) at any time ${t}$ satisfies:
\begin{equation}
  \label{eq:mutFAST}  
\mathbf{\mu}(t) = \mathbf{\mu}_{\text{base}} + \Delta \mathbf{\mu}(t)
\end{equation}
where \(\mathbf{\mu}_{\text{base}} = [x_{\text{base}}, y_{\text{base}}, z_{\text{base}}]^T\) denotes the static 3D Gaussian center, and \(\Delta \mathbf{\mu}(t)\) represents the displacement calculated via Fourier series to capture dynamic variations. For ${x}$, ${y}$, ${z}$ directions (independently calculated),  \(\Delta \mathbf{\mu}(t)\) includes a constant term and a Fourier series term, expressed as:
\begin{equation}
\label{eq:deltaxyzFAST}
\begin{split}
\Delta x(t) &= A_{x,0} + \sum_{k=1}^{L} \left[ A_{x,k} \cdot \sin(2\pi kt) + B_{x,k} \cdot \cos(2\pi kt) \right] \\
\Delta y(t) &= A_{y,0} + \sum_{k=1}^{L} \left[ A_{y,k} \cdot \sin(2\pi kt) + B_{y,k} \cdot \cos(2\pi kt) \right] \\
\Delta z(t) &= A_{z,0} + \sum_{k=1}^{L} \left[ A_{z,k} \cdot \sin(2\pi kt) + B_{z,k} \cdot \cos(2\pi kt) \right]
\end{split}
\end{equation}
where the coefficients ${A}$ and ${B}$ represent the ${k}$-th order coefficients for the displacement 
formulas in the ${x}$, ${y}$, ${z}$ directions respectively. In our implementation, we use ${L}$ = 5 as we find it a good balance between representation precision and model efficiency.

We employ variations in opacity to represent the appearance or disappearance of a Gaussian at a given time ${t}$. For a spacetime point ($\mathbf{x}$, ${t}$), the opacity of an Gaussian is
\begin{equation}
  \label{eq:opacity}  
\alpha(t)=\sigma(t)\exp\left(-\frac{1}{2}(\mathbf{x}-\mu(t))^{T} \Sigma(t)^{-1}(\mathbf{x}-\mu(t))\right)
\end{equation}
where $\sigma(t)$ is temporal opacity, $\Sigma(t)$ is the time-dependent covariance and $\mu(t)$ is the instantaneous position in Eq.(\ref{eq:mutFAST}).The detail about each of the components are the following:

Inspired by \cite{chen2023neurbf,kerbl20233d,li2024spacetime} that use radial basis functions for approximating spatial signals, we model the temporal opacity $\sigma(t)$ of a Gaussian using a 1D radial basis function:
\begin{equation}
  \label{eq:temporal-opacity}  
\sigma(t) = \sigma^s \exp\left(-s^\tau |t - \mu^\tau|^2\right)
\end{equation}
where $\mu^\tau$ denotes the temporal center (peak visibility time), $s^\tau$ controls the temporal scaling (duration of visibility), and $\sigma^s$ represents the spatial opacity independent of time. This formulation allows each Gaussian to appear and disappear smoothly over time.

Following \cite{4dgsRealtime,li2024spacetime}, we parameterize the rotation matrix ${R}(t)$ using a real-valued quaternion ${q}(t)$, represented by the instantaneous position in Eq.(\ref{eq:mutFAST}) of time:
\begin{equation}
  \label{eq:temporal-quaternion}  
{q}(t) = \sum_{k=0}^{n_q} \mathbf{c}_{k} (t - \mu^\tau)^k
\end{equation}
where ${\mathbf{c}_{k}}$ are polynomial coefficients and $n_q=1$ in our implementation. Notably, the scaling matrix ${S}$ remains time-independent, as temporal variation did not improve rendering quality in experiments\cite{li2024spacetime}.

To ensure model compactness, we follow the same approach as \cite{li2024spacetime} by employing feature vectors rather than spherical harmonic (SH) coefficients to store the RGB color of each Gaussian. The feature splatting process is analogous to standard Gaussian splatting\cite{kerbl20233d}, 
with the key difference that the RGB color $c_i$ in Eq.(\ref{eq:colorrendering}) is replaced by a feature vector $f_i(t)$. 

% --- 插入开始 ---

\begin{table}[b] % [htb] 表示尝试放在 这里(h)、页顶(t) 或 页底(b)
 \centering
 \scriptsize
 \caption{Quantitative comparison on generalization and efficiency. We randomly selected two views from the \textit{02\_Flames} scene in Google Immersive for additional experiments. The efficiency experiment was conducted at viewpoint cam0010.}
 \label{tab:generalization and efficiency}
 % 使用 resizebox 确保表格宽度刚好填满单栏，不会溢出
 \resizebox{\linewidth}{!}{
 \begin{tabular}{lccccc}
 \toprule
 \multirow{2}{*}{\textbf{Method}} & \multicolumn{2}{c}{\textbf{PSNR} (dB) $\uparrow$} & \textbf{Speed} & \textbf{Storage} & \textbf{Training} \\
 \cmidrule(lr){2-3}
 & cam0010 & cam0037 & (FPS) $\uparrow$ & (MB) $\downarrow$ & Time $\downarrow$ \\
 \midrule
 Hybrid 3D-4D\cite{oh2025hybrid} & 28.51 & 29.61 & 137 & 373 & \textbf{31m27s} \\
 STGFS\cite{li2024spacetime}        & 27.97 & 27.64 & \textbf{154} & \textbf{248} & 34m11s \\
 \textbf{Ours}& \textbf{30.56} & \textbf{31.54} & 146 & 363 & 40m34s \\
 \bottomrule
 \end{tabular}
 }

\end{table}

% --- 插入结束 ---

\subsection{\textit{Motion Regularization}}
\label{ssec:subhead}
To control the parameter complexity of the Fourier approximation, suppress high-frequency jitter while preserving motion details, and mitigate overfitting to noise in the training data, we propose a motion regularization loss function defined as follows:
\begin{equation}
  \label{eq:motion-regularization}  
\mathcal{L}_{M-reg} = \sum_{i=1}^{L}\sum_{j=0}^{2}\left( i^{2}\cdot(\|{w}_{j,2i-1}\|_{2}+\|{w}_{j,2i}\|_{2})\right)
\end{equation}
where \( j \in \{0,1,2\} \) denotes \( x/y/z \) directions, and \( w_{j,2i-1}/w_{j,2i} \) are sin/cos coefficients for the \( i \)-th frequency. The weight \( i^2 \) implements a low-pass filter: higher frequencies (larger \( i \)) are penalized more heavily, suppressing jitter without blurring low-frequency motion. 

\begin{table*}[!t]
  \centering
  \scriptsize
  \caption{Quantitative comparison on N3V dataset across multiple scenes: \textit{Coffee Martini}, \textit{Cook Spinach}, \textit{Flame Steak}, \textit{Sear Steak} — each with a duration of 10 seconds. The \colorbox{orangehighlighter}{best} and \colorbox{yellowhighlighter}{second best} scores among competing methods are highlighted (Exclude NeRF-based methods\cite{attal2023hyperreel,kplanes,song2023nerfplayer}).}
  \label{tab:quantitative_results_full}
  \setlength{\tabcolsep}{1.8pt}
  \renewcommand{\arraystretch}{1.0}
  \begin{tabular}{l*{4}{cccc}}
    \toprule
    \multirow{2}{*}{\textbf{Method}} & \multicolumn{4}{c}{\textbf{PSNR} $\uparrow$} & \multicolumn{4}{c}{\textbf{DSSIM1} $\downarrow$} & \multicolumn{4}{c}{\textbf{DSSIM2} $\downarrow$} & \multicolumn{4}{c}{\textbf{LPIPSAlex} $\downarrow$} \\
    \cmidrule(lr){2-5} \cmidrule(lr){6-9} \cmidrule(lr){10-13} \cmidrule(lr){14-17}
    & \textit{Sear} & \textit{Flame} & \textit{Cook} & \textit{Coffee} & \textit{Sear} & \textit{Flame} & \textit{Cook} & \textit{Coffee} & \textit{Sear} & \textit{Flame} & \textit{Cook} & \textit{Coffee} & \textit{Sear} & \textit{Flame} & \textit{Cook} & \textit{Coffee} \\
    & \textit{Steak} & \textit{Steak} & \textit{Spinach} & \textit{Martini} & \textit{Steak} & \textit{Steak} & \textit{Spinach} & \textit{Martini} & \textit{Steak} & \textit{Steak} & \textit{Spinach} & \textit{Martini} & \textit{Steak} & \textit{Steak} & \textit{Spinach} & \textit{Martini} \\
    \midrule
    NeRFPlayer\cite{song2023nerfplayer}       & 29.13                               & 31.93                               & 30.56                               & 31.53                               & 0.0460                               & 0.0250                               & 0.0355                               & 0.0245                               & —                                    & —                                    & —                                    & —                                    & 0.0138                               & 0.0880                               & 0.1130                               & 0.0850 \\
    HyperReel\cite{attal2023hyperreel}        & 32.57                               & 32.20                               & 32.30                               & 28.37                               & 0.0240                               & 0.0255                               & 0.0295                               & 0.0540                               & —                                    & —                                    & —                                    & —                                    & 0.0770                               & 0.0780                               & 0.0890                               & 0.1270 \\
    K-Planes\cite{kplanes}                    & 32.52                               & 32.38                               & 31.82                               & 29.99                               & —                                    & —                                    & —                                    & —                                    & 0.0132                               & 0.0153                               & 0.0169                               & 0.0162                               & —                                    & —                                    & —                                    & — \\
    \midrule
    Dynamic 3DGS\cite{luiten2024Dynamic3D}    & 33.68                               & 33.24                               & 32.97                               & 26.49                               & 0.0224                               & 0.0233                               & 0.0263                               & 0.0557                               & 0.0105                               & 0.0113                               & 0.0129                               & 0.0332                               & 0.0790                               & 0.0790                               & 0.0870                               & 0.1390 \\
    4DGS\cite{4dgsdeform}                     & 33.01                               & 31.83                               & 33.06                               & 27.88                               & 0.0237                               & 0.0248                               & 0.0267                               & 0.0470                               & 0.0125                               & 0.0137                               & 0.0142                               & 0.0284                               & 0.0416                               & 0.0418                               & 0.0519                               & 0.8550 \\
    4DGS\cite{4dgsRealtime}                   & 33.44                               & 33.19                               & 32.73                               & 27.98                               & \colorbox{yellowhighlighter}{0.0204} & \colorbox{yellowhighlighter}{0.0204} & 0.0245                               & 0.0435                               & 0.0105                               & \colorbox{yellowhighlighter}{0.0106} & 0.0133                               & 0.0265                               & 0.0411                               & 0.0389                               & 0.0489                               & 0.8470 \\
    4DGS-1K\cite{yuan20251000+}               & 33.60                               & 33.25                               & 33.06                               & 28.54                               & 0.0396                               & 0.0419                               & 0.0460                               & 0.0834                               & —                                    & —                                    & —                                    & —                                    & 0.0402                               & 0.0421                               & 0.0467                               & 0.0744 \\
    Ex4DGS\cite{lee2024fully}                 & 33.69                               & \colorbox{orangehighlighter}{33.91} & 33.23                               & 28.79                               & 0.0410                               & 0.0440                               & 0.0530                               & 0.0850                               & 0.0210                               & 0.0200                               & 0.0240                               & 0.0490                               & 0.0350                               & 0.0340                               & 0.0420                               & 0.0700 \\
    Hybrid 3D-4D\cite{oh2025hybrid}           & \colorbox{yellowhighlighter}{34.45} & \colorbox{yellowhighlighter}{33.79} & \colorbox{yellowhighlighter}{33.42} & \colorbox{yellowhighlighter}{28.86} & 0.0225                               & 0.0217                               & 0.0362                               & 0.0541                               & 0.0120                               & 0.0117                               & 0.0228                               & \colorbox{yellowhighlighter}{0.0241} & 0.0317                               & \colorbox{orangehighlighter}{0.0294} & \colorbox{yellowhighlighter}{0.0349} & 0.0986 \\
    MEGA\cite{zhang2024mega}                  & 33.67                               & 32.27                               & 33.08                               & 27.84                               & \colorbox{orangehighlighter}{0.0200} & 0.0242                               & \colorbox{yellowhighlighter}{0.0230} & 0.0440                               & 0.0103                               & 0.0129                               & \colorbox{yellowhighlighter}{0.0125} & 0.0270                               & 0.0403                               & 0.0538                               & 0.0471                               & 0.0770 \\
    STGFS\cite{li2024spacetime}               & 33.40                               & 33.59                               & 33.30                               & 28.55                               & 0.0351                               & 0.0350                               & 0.0425                               & \colorbox{yellowhighlighter}{0.0418} & \colorbox{yellowhighlighter}{0.0064} & \colorbox{orangehighlighter}{0.0053} & \colorbox{orangehighlighter}{0.0046} & 0.0253                               & \colorbox{yellowhighlighter}{0.0309} & \colorbox{yellowhighlighter}{0.0308} & 0.0358                               & \colorbox{yellowhighlighter}{0.0692} \\
    Ours                                      & \colorbox{orangehighlighter}{34.65} & 33.09                               & \colorbox{orangehighlighter}{33.49} & \colorbox{orangehighlighter}{29.98} & 0.0298                               & \colorbox{orangehighlighter}{0.0195} & \colorbox{orangehighlighter}{0.0211} & \colorbox{orangehighlighter}{0.0367} & \colorbox{orangehighlighter}{0.0022} & 0.0165                               & 0.0131                               & \colorbox{orangehighlighter}{0.0167} & \colorbox{orangehighlighter}{0.0306} & 0.0399                               & \colorbox{orangehighlighter}{0.0338} & \colorbox{orangehighlighter}{0.0583} \\
    \bottomrule
  \end{tabular}
\end{table*}

Then for the regularizing iterations, the total loss function is taken to be
\begin{equation}
  \label{eq:loss}  
\mathcal{L} = \mathcal{L}_1 + \mathcal{L}_{DSSIM} + \lambda_{M-reg}\mathcal{L}_{M-reg}
\end{equation}
where $\mathcal{L}_1$ (pixel-wise loss) and $\mathcal{L}_{DSSIM}$ (structural similarity) ensure photorealism, and $\lambda_{M-reg}$ denotes adjustable weights that can be customized according to specific scenarios.

\begin{table}[h]
 \centering
 % 1. 显式设置字体大小，确保和上面的表格看起来差不多大
 \scriptsize
 % 2. 关键：加大列与列之间的间距。
 % 因为这个表格内容少，加大间距可以让表格看起来更舒展，铺满单栏，而不会把字变大。
 % 你可以调整这个数值（比如 8pt, 10pt, 12pt）来控制表格的总宽度。
 \setlength{\tabcolsep}{12pt}
 \caption{Ablation studies on Fourier order ($L$) and regularization weight ($\lambda$). Evaluated on the \textit{Cook Spinach} scene. We adopt $L=5$ and $\lambda=0.1$ as the default settings.}
 \label{tab:ablation_study}
 % 3. 去掉 \resizebox，直接放 tabular
 \begin{tabular}{cccl}
 \toprule
 \textbf{Experiment} & \textbf{Value} & \textbf{PSNR} & \textbf{Observation} \\
 \midrule
 % 第一组：Fourier Order (L)
 \multirow{4}{*}{\shortstack{Fourier\\Order ($L$)}}
 & 1 & 28.45 & Low fidelity \\
 & 3 & 31.12 & Improved \\
 & \textbf{5 (Ours)} & 33.62 & \textbf{Optimal Balance} \\
 & 7 & \textbf{33.79} & +19\% Train Time \\
 \midrule
 % 第二组：Regularization Weight (lambda)
 \multirow{4}{*}{\shortstack{Reg. Weight\\($\lambda$)}}
 & 0 & 30.12 & High-freq Jitter \\
 & 0.01 & 30.95 & Minor Jitter \\
 & \textbf{0.1 (Ours)} & \textbf{33.62} & \textbf{Best Quality} \\
 & 1.0 & 29.40 & Over-smoothed \\
 \bottomrule
 \end{tabular}
\end{table}

%第三章：实验和讨论
\section{Experiments}
\label{sec:majhead}
\subsection{\textit{Datasets}}
\label{ssec:subhead}
We evaluate our proposed method on the N3V dataset\cite{li2022neural}, which includes 6 multi-view video sequences (18–21 cameras, 2704×2028 resolution).
In addition, we also conducted experiments on Google Immersive\cite{broxton2020immersive} to validate the generalization of our model on different datasets.
Consistent with previous work, we reserve cam00 as the test view and use the remaining cameras for training on N3V dataset. For Google Immersive, we use the first $80\%$ frames for training and use the last $20\%$ frames for testing.
In accordance with common experimental practice, we adopt the same settings for all previous methods during the experiment process \cite{4dgsdeform,4dgsRealtime,lee2024fully,li2024spacetime,zhang2024mega,yuan20251000+,oh2025hybrid,luiten2024Dynamic3D} to ensure a fair comparison.

\subsection{\textit{Results and Analysis}}
\label{ssec:subhead}
As shown in Table \ref{tab:quantitative_results_full}, our method demonstrates competitive or state-of-the-art performance across four challenging dynamic scenes compared to existing Gaussian Splatting-based approaches. It performs particularly well in scenarios involving complex motion patterns.

In terms of reconstruction fidelity (PSNR), our method achieves the highest scores on \textit{Sear Steak} $(34.65)$ and \textit{Cook Spinach} $(33.49)$, and ranks second on \textit{Coffee Martini} $(29.98)$. This consistent performance across diverse scene types underscores the robustness of our frequency-aware motion modeling, especially under significant non-rigid deformations and fast motion.

On structural similarity metrics, for fair comparison, we group existing methods’ DSSIM results into DSSIM1 (range 1.0) and DSSIM2 (range 2.0). Our method excels in DSSIM2 on \textit{Sear Steak}, significantly outperforming other methods. It also achieves the best DSSIM1 results on \textit{Flame Steak} $(0.0195)$ and \text{Cook Spinach} $(0.0211)$, confirming its ability to preserve structural consistency.

For perceptual quality (LPIPSAlex), our method obtains the best results on \textit{Cook Spinach} $(0.0338)$ and \textit{Coffee Martini} $(0.0583)$, and ranks second on \textit{Sear Steak} $(0.0306)$, indicating improved visual realism beyond numerical metrics.

Also shown in Table \ref{tab:generalization and efficiency}, our model still demonstrates excellent performance even when the test view is randomly selected. Moreover, the slight increase in cost is fully offset by the significant improvement in quality. Notably, our approach maintains strong performance across all scenes without noticeable degradation, demonstrating the generalization capacity of the frequency-aware representation.
% %图4:消融实验
% \begin{figure}

% \begin{minipage}[b]{1.0\linewidth}
%   \centering
%   \centerline{\includegraphics[width=8cm]{xiaorong.pdf}}
% %  \vspace{2.0cm}
%   \centerline{(a) Training View}\medskip
% \end{minipage}
% \begin{minipage}[b]{1.0\linewidth}
%   \centering
%   \centerline{\includegraphics[width=5cm]{xiaorong2.pdf}}
% %  \vspace{1.5cm}
%   \centerline{(b) Novel View}\medskip
% \end{minipage}
% \caption{$\textbf{Ablation with motion regularization.}$ With motion regularization, The rendering results of rapidly moving objects (such as the spatula in cook spinach) are more realistic.}
% \label{fig:Ablation}
% %
% \end{figure}

\subsection{\textit{Ablation Study}}
\label{ssec:subhead}
\begin{figure}[ht]
    \centering
    \includegraphics[width=1.0\linewidth]{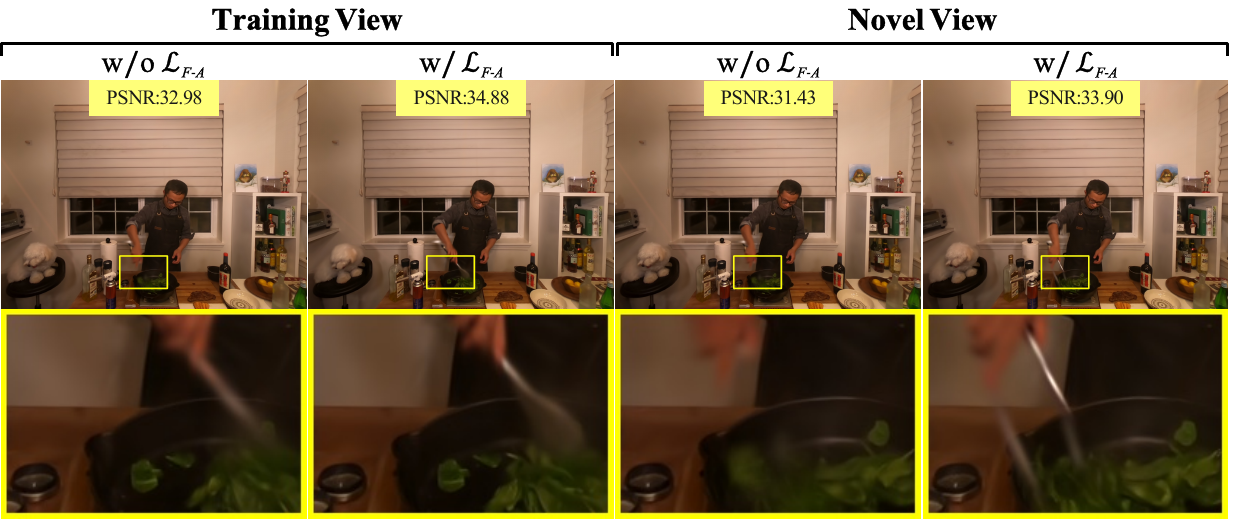}
    \caption{$\textbf{Ablation with motion regularization.}$ With motion regularization, the rendering results of rapidly moving objects (such as the spatula in cook spinach) are more realistic.
} 
    \label{fig: res-ours}
\end{figure}

To validate the effectiveness of motion regularization, we conducted an ablation study. As shown in Fig.\ref{fig: res-ours} and Table \ref{tab:ablation_study}: motion regularization can significantly improve the rendering quality of rapidly moving objects, and the best balance between performance and efficiency is achieved when the Fourier Order ${L}$ = 5 and ${\lambda}$ = 0.1 are set appropriately.

% 在需要插入图像的位置使用此代码

\section{Conclusion}
\label{sec:print}

We propose FAST-GS, a frequency-aware 4D Gaussian Splatting method for dynamic novel view synthesis. By replacing polynomial motion modeling with fourier decomposition, we address 4DGS’s inability to fit complex motion and maintain long-term stability. Our motion-aware regularization further balances noise suppression and detail preservation. Experiments on N3V and Google Immersive confirm that FAST-GS enhances rendering quality in complex scenes while retaining real-time performance.

% Although our representation enables high-fidelity rendering, the current method is primarily designed for multi-view video inputs. Extending the framework to monocular settings—potentially through advanced regularization techniques 
% or generative priors—constitutes a promising direction for future research.

\vfill\pagebreak

\bibliographystyle{IEEEbib}
\bibliography{strings, refs}

\end{document}